\documentclass[conference]{IEEEtran} \IEEEoverridecommandlockouts
\usepackage{cite}
\usepackage{amsmath,amssymb,amsfonts}
\usepackage{algpseudocode}
\usepackage{multirow}
\usepackage{graphicx}
\usepackage{graphics}
\usepackage{textcomp}
\usepackage{caption}
\usepackage{subcaption}
\usepackage{xcolor}

\usepackage{comment}

\def\BibTeX{{\rm B\kern-.05em{\sc i\kern-.025em b}\kern-.08em
    T\kern-.1667em\lower.7ex\hbox{E}\kern-.125emX}}
\title{A Review of Machine Learning Methods Applied to Video Analysis
  Systems}

\author{
  Marios S. Pattichis$^\ast$, Venkatesh Jatla$^\ast$, and
  Alvaro E. ulloa Cerna$^\ddagger$\\
  $^\ast$Department of Electrical and Computer Engineering\\ The
  University of New Mexico, Albuquerque, NM, USA.
  Email: \{pattichi, venkatesh369\}@unm.edu\\
  $^\ddagger$Pontificia Universidad Catolica del Peru, Lima, Peru
  Email: alvarouc@gmail.com\\ }
\begin{document}
\maketitle

\begin{abstract}
  The paper provides a survey of the development of machine-learning techniques
      for video analysis.
  The survey provides a summary of the most popular
      deep learning methods used for human activity recognition.
  We discuss how popular architectures perform
      on standard datasets and highlight
      the differences from  
      real-life datasets
      dominated by multiple activities performed
      by multiple participants over long periods.
  For real-life datasets, we describe the use
      of low-parameter models (with 200X or 
      1,000X fewer parameters)
      that are trained to detect a single activity
      after the relevant objects have been
      successfully detected.

   Our survey then turns to a summary of machine learning
       methods that are specifically developed for working
       with a small number of labeled video samples.
   Our goal here is to describe modern techniques that
       are specifically designed so as to minimize
       the amount of ground truth that is needed for
       training and testing video analysis systems.
   We provide summaries of the development of
       self-supervised learning, 
       semi-supervised learning,
       active learning, and 
       zero-shot learning for applications in 
       video analysis.
   For each method, we provide representative examples.    
\end{abstract}


\section{Introduction}
Video analysis has been greatly impacted by the
    recent advances in deep learning methods.
Deep learning methods are increasingly applied
    in all areas of video analysis.
The majority of video analysis are trained
    using supervised learning, where
    training a large deep-learning
    system requires a large dataset.
There are several challenges associated
    with this standard paradigm.
First, labeling a large number of
    video samples is very time 
    consuming.
Second, training an end-to-end 
    system on a large number of samples
    can be very slow, requiring significant
    computational resources.

The majority of current video
     analysis systems are trained on a
     relatively small number of samples over
     a limited number of human activities.
The UCF101\cite{soomro2012ucf101} is one of the most popular action
     recognition datasets.
UCF101 contains 101 action classes with over 13,000 video
     samples for a total of 27 hours.
Video segments average 7.21 seconds at 25 FPS at a resolution of 320 × 240 pixels.
HMDB51\cite{kuehne2011hmdb} is another popular action recognition 
      dataset.
HMDB51 contains 51 action classes with around 7,000 samples, mostly extracted from movies. 
Video segments are between 2 to 5 seconds at 30 FPS rescaled to a height of 240 pixels.
In comparison, the original ImageNet dataset contained about 1.3 million images
       with 1,000 categories.
Unfortunately, while much larger video datasets have become available,
    it is computationally very expensive to train on them.

\begin{figure}
\includegraphics[width=\linewidth]{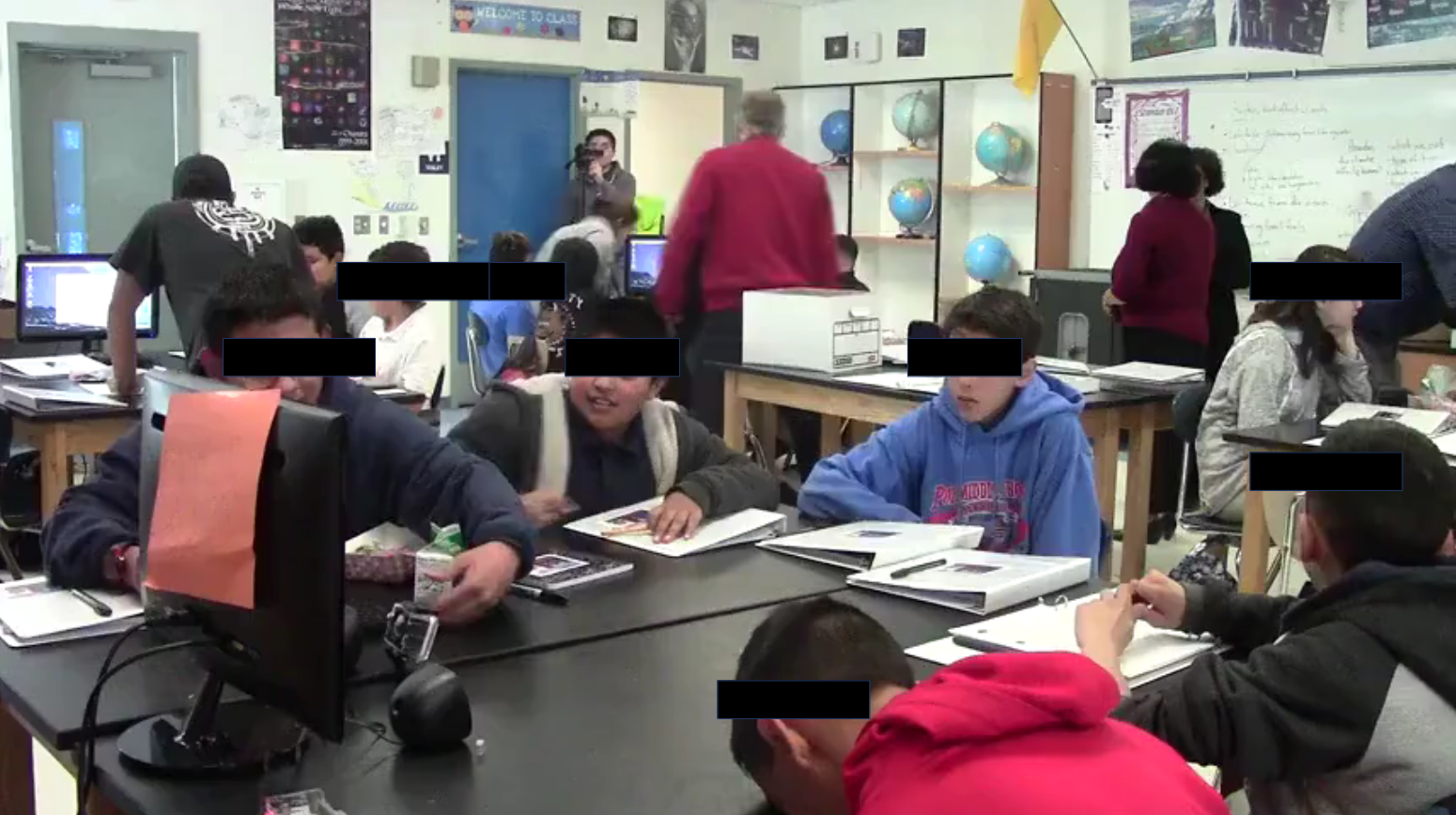}
\caption{Example of complex video content from AOLME\cite{aolme} video
data set. The dataset cotains a total of 2,218 hours
of transcodedd video at $858\times480$ resolution and $30$ Frames Per Second (FPS).}
\label{fig:aolme_sample}
\end{figure}

To appreciate the complexity of processing real-life video datasets,
    we present an example in Fig. \ref{fig:aolme_sample}.
The image includes several participants
    appearing at different angles, performing
    a variety of different activities.
We can identify objects and humans associated
    with specific activities
    by analyzing a select number of video frames.
We then need to identify the activity associated
    with each detected object and track the activity
    throughout the video segment.

We note that there are significant differences
    between the real-life example of Fig. \ref{fig:aolme_sample}.
    and the standard datasets used for video action recognition.
First, we note that we have multiple activities performed
     by different people.
Second, we note that these activities are occurring at 
     many different scales with significant partial and total
     occlusions.
Third, we have people entering or leaving the scene.
Fourth, in terms of duration, the video sessions
     range from 1 hour to 90 minutes.
Fifth, unlike the standard datasets, these real-life
     datasets exhibit a relatively small number of
     actual video activities of interest.     

In our survey, we will provide an overview of 
    the most popular human activity recognition
    systems used for standard datasets.
Furthermore, we will also summarize
    our own efforts to develop
    real-life video activity recognition
    systems using low-parameter models
    that are separately trained for each activity.

We also describe different
   learning methods aimed at minimizing the 
   required number of labeled samples.
Our goal here is to minimize the 
    amount of effort required to 
    develop ground truth on
    video datasets.
We provide an overview of methods associated 
    with self-supervised learning, 
   semi-supervised learning,
   active learning, and 
   zero-shot learning.
For each learning method,
   we provide the relevant definitions
   and specific examples from video analysis.

The rest of the paper is broken into three sections.
In section \ref{sec:human-activity},
    we summarize some of the most popular
    methods for human activity recognition.
We also present our 
    development of low-parameter models
    for real-life videos in this section.
In section \ref{sec:learning},
    we describe different learning methods
    for training video analysis systems with a limited
    number of labeled samples.
We provide concluding remarks in section \ref{sec:conclusion}.    

\section{Human Activity Recognition Systems}\label{sec:human-activity}
We provide an overview of commonly used  
    Human Activity Recognition (HAR) systems as summarized in Table
    \ref{tab:har-frameworks}. 
In what follows, we will provide a brief description
    for each one of these popular methods.
At the end of the section, we provide
     a summary of our development
     of low-parameter models for processing
     real-life video datasets.

\begin{table*}[!t]
  \caption{Summary of Human Activity Recognition (HAR) frameworks.}
  \label{tab:har-frameworks}
  \begin{tabular}{ p{2cm} p{12.25cm}}
    \hline
    \textbf{Dataset}                                                       & \textbf{Summary} \\
    \hline
    \hline
    TSN \cite{tsn} \newline (2019) & 
        $-$ Has three paths: RGB, frame difference, and optical flow. \newline
        $-$ Video split into multiple segments, each contributing a class score. \newline
        $-$ Aggregating the scores determines the final class. \newline
        $-$ 24M parameters. \newline
        $-$ Method achieved an accuracy of 70\% on Kinetics-400 \cite{kinetics} dataset. \newline
    \\
    TSM \cite{tsm} \newline (2019) &
        $-$ High efficiency and high performance \newline
        $-$ Method achieved 3D-CNN performance maintaining 2D-CNN complexity \newline
    \\
    Slowfast \cite{slowfast}\newline (2019) & 
        $-$ Two pathways: spatial (slow) and temporal (fast).\newline
        $-$ The paths can use 2D or 3D-CNNs. \newline
        $-$ The paper uses ResNet \cite{resnet2016} to design the pathways. \newline
        $-$ 32M parameters. \newline
        $-$ Method achieved an accuracy of 74\% on Kinetics-400 \cite{kinetics} dataset. \newline
    \\
    I3D \cite{i3d} \newline (2017) & 
        $-$ Proposed building and initializing 3D-CNNs by “inflating” famous 2D-CNN architectures. \newline
        $-$ Paper inflates Inception network. \newline
        $-$ 27M parameters \newline
        $-$ Method achieved an accuracy of 72\% on Kinetics-400 \cite{kinetics} dataset.\newline
    \\
    LT-HAQ \cite{jatla2021long} \newline (2021) &
    $-$ Proposed low-parameter models to classify typing and writing videos. \newline
    $-$ Achieves similar performance as other high complexity
    models with 1200x to 1500x less parameters. \newline
    $-$ 19K parametrs
    $-$ The model uses only 350 MB of video memory. \newline
    $-$ The spatio-temporal regions are determined using object detection and tracking frameworks.\\
    
    \hline
  \end{tabular}
\end{table*}

\textbf{Temporal Segment Network (TSN):}
TSN \cite{tsn} attempts to model long-term 
    activities. 
It uses fixed sparse temporal sampling.
TSN achieved good performance over 
   HMDB51 (69.4\%) and UCF101 (94.2\%). 
The use of sparse sampling performs well
   on video activities with unique temporal patterns.

\textbf{Two-Stream Inflated 3D ConvNet (I3D):} 
The goal of I3D is to adopt state-of-the-art 
    image classification architectures (e.g., Inception), 
    and inflate the filters and pooling kernels into 3D
    for analyzing digital videos \cite{i3d}. 
Thus, I3D builds robust representations derived from 2D
    images. 
To enhance its performance, I3D incorporates two input
    streams: RGB and optical flow, both initialized from the weights of
    the 2D networks and expanded to 3D. 
I3D gave 80.9\% accuracy on HMDB-51 and 98.0\% on UFC-101.

\textbf{Temporal Shift Module (TSMs):}
TSM \cite{tsm} is a highly efficient and high-performance 
    model that achieves 3D CNN-level performance 
    while maintaining the complexity of a 2D CNN. 
By moving a portion of the channels along the temporal axis,
    TSM facilitates communication between neighboring frames and enables
    efficient temporal modeling. 
In offline tests, TSM achieved impressive results: 
    74.1\% accuracy on Kinetics, 95.9\% on UCF101, and 73.5\% on HMDB51. 
Online, for real-time applications, TSM achieved 74.3\%, 95.5\%, and 73.6\% on 
   the same datasets respectively. 

\textbf{SlowFast:}
SlowFast \cite{slowfast} is a video analysis model
       that comprises a Slow and a Fast pathway. 
The Slow pathway operates at a lower frame rate and captures 
       spatial semantics, while the Fast pathway 
       operates at a higher frame rate and captures motion
       at a finer temporal resolution.
SlowFast models have demonstrated strong performance in both action
       classification and detection in video, with significant improvements
       attributed to the SlowFast concept. 
The Slow pathway in a SlowFast network is designed to have 
       a low frame rate and lower temporal resolution, 
       while the Fast pathway has a high frame rate and greater
       temporal resolution.
Overall, the SlowFast model architecture provides a powerful and
       effective means of capturing spatio-temporal features from video, with
       the Slow and Fast pathways working together to achieve impressive
       results in video analysis tasks. 

\textbf{Low-parameter models for real-life datasets:}
Instead of developing universal classifiers for detecting
    all activities, we examine an alternative approach that
    trains a unique classifier for each activity.
The basic approach was first demonstrated in \cite{ulloa2021deep}.
In \cite{ulloa2021deep}, the authors
    trained a 3D CNN with just 20K parameters on
    812,278 echocardiographic videos from 34,362 individuals 
    to predict one-year all-cause mortality.
The model performed very well.
Here, we note that single-activity classifiers
    do not need to be adaptations of 
    universal classifiers that attempt
    to classify all possible activities.
Instead, they only need to learn to recognize
    a single activity.

We have adopted this approach for human activity
    recognition of our real-life classroom videos.
First, we decouple video activity recognition from
    the need to localize the activity.
We used standard object detection methods
     to locate humans (YOLO \cite{redmon2018yolov3},
     Faster-RCNN \cite{ren2015faster}, or other representations),
     and Arcface for face detection \cite{deng2019arcface}.
We then track the objects through time to generate proposals
     of possible activities (e.g., see \cite{teeparthi2021fast}).
To recognize the activity within the proposed video segment,
      we use low-parameter 3D-CNN models
      (e.g., \cite{jatla2021long, shi2021talking}). 
For typing and writing activities,
    the low-parameter 3D CNN achieved an 80\% accuracy rate
    in detection, comparable to the performance achieved by
    TSN, SlowFast, and I3D, but with 200x to 1500x fewer parameters.

\section{Learning with A Limited Number of Labeled Samples}\label{sec:learning}
We examine four different learning paradigms
   that have been used to train video datasets
   with a limited number of labeled samples.
We begin with self-supervised learning where
   the training is performed without any user-provided
   labels.
We then cover semi-supervised learning where
   our goal is to spread a limited number of labels
   to a wider set of unlabeled samples.
We tackle the problem of minimizing the number
   of labeled samples in active learning.
Finally, in zero-shot learning, we discuss
   methods that can learn new video activities
   using pre-trained systems, without the need
   for newly labeled samples.

\subsection{Self-supervised learning}
Self-supervised learning refers to the process of learning
    models from unlabeled data.
A standard approach is to predict portions of a video
    from the rest of the video.
The idea here is that we can use self-supervised
    learning on a large unlabeled dataset and
    then use the trained model on a task where
    the small number of labels do not allow standard
    supervised training.
For the Signal Processing community, self-supervised learning
     sounds similar to video processing in the compressed domain.
The motivation here is different.
Here, the motivation is to use a large
     unlabeled dataset to train a deep learning model
     with a large number of parameters that can
     be proven useful for a simpler task.
There is a lot of activity in this area.
We refer to \cite{schiappa2023self} for a recent survey.

In \cite{schiappa2023self},  progress in the field 
     is measured in terms of accuracy performance achieved on
     the UCF101 and HMDB51 video activity datasets.
Here, it is noted that the use of multimodal data
     (video+audio+text) provides the best results.
The authors organize the literature into methods
     based on pretext tasks, generative learning,
     contrastive learning, and cross-modal agreement.
An example of a pretext task is to rotate the video
     and train the network to predict the rotation angle.
Another example includes changing the speed of the video and      
      predicting its changed speed.
In generative learning, videos can be generated using
      GANs or predict masked tokens from the rest of them.
In contrastive learning, the goal is to develop methods 
      that differentiate between positive and negative samples.
Several video augmentation methods fall under contrastive learning.      
      
\subsection{Semi-supervised learning}
In semi-supervised learning, our goal is
   combine supervised and unsupervised learning methods
   to generate better classifiers.
Typically, we assume that we are given
   a small number of labeled samples and
   a large number of unlabeled samples.

We provide a simplified algorithm
   of semi-supervised learning in
   Fig. \ref{fig:semi-supervised}.
Initially, we perform standard supervised learning
    to produce an initial classifier over a limited
    number of labeled samples.
We then employ an unsupervised
    technique to spread the current
    labels over a large number of unlabeled samples.
For the unsupervised technique,
     we can look at the nearest neighbors of
     labeled samples, a classifier method with
     high probability, or a combination of 
     measures based on classifier prediction and
     sample similarity.
As an example, a classifier that predicts
     a specific class with a high probability 
     is expected to correspond 
     to a high-confidence classification.
Similarly, when two samples are similar,
     they are expected to belong to the same class.
Clearly, the success of semi-supervised
      learning depends on our ability to generate
      correct new labels.
Here, a possible variation is to apply soft labeling
       and then use expectation maximization to 
       relabel the samples.

When using standard classifiers,
     it is important to calibrate
     them prior to using their outputs    
     to label new samples.
Briefly, over the range of zero to 1, 
    calibration involves a process of adjusting the 
    classifier parameters to ensure that the mean prediction probability 
    corresponds to the predicted positive fraction
    (see \cite{selfcal}).
Unfortunately, classifier calibration may not always
    be possible for more complex networks.

In \cite{jing2021videossl}, the authors demonstrate 
     excellent performance on the UCF101, HMDB51, and Kinetics datasets using a 
     semi-supervised approach.
When using a small percentage of the original labeled samples,
     the authors showed that their semi-supervised technique
     strongly outperformed fully supervised training over
     the same percentage.
After initial training on a small labeled dataset,
      the 3D network was trained on a combination of three cross-entropy measures
      computed over the labeled video samples, pseudo-labels over the unlabeled data,
      and the soft loss based on appearance.
For the soft loss component, the authors compared the 
      outputs of a 2D CNN meant to capture appearance over sampled frames
      with a 3D CNN that uses a reshaped output to match the 2D CNN output.
      
\begin{figure}[!b]
  \begin{algorithmic}[1]
    \State \textbf{Sample and provide GT} from independent video sessions
    \State \textbf{Train} initial model
    \While {no new samples can be generated}
      \State \textbf{Select unlabeled samples} based on \Statex $\qquad\quad$ similarity to labeled samples
      \State \textbf{Use} classifier pseudo-labels or
       propagate labels to
       \Statex $\qquad\quad$ unlabeled samples
      \State \textbf{Train} model with larger dataset
    \EndWhile
  \end{algorithmic}
  \caption{A simple algorithmic framework for semisupervised learning}
  \label{fig:semi-supervised}
\end{figure}

Semi-supervised learning, on the other hand, sits between supervised and unsupervised learning. It uses a small amount of labeled data along with a larger pool of unlabeled data. The idea is to leverage the labeled data to guide the learning process and help the model make better use of the unlabeled data. Techniques in semi-supervised learning include methods like self-training \cite{amini2022self}, where a model is initially trained with a small labeled dataset and then used to label the unlabeled data, and consistency regularization\cite{zhang2019consistency}, where the model is encouraged to produce consistent predictions when small perturbations are applied to the data.

\subsection{Active learning} 
Active learning aims to reduce the amount of 
    required data annotation through sample selection.
The goal here is to select 
    samples that can substantially improve the performance 
    of the classifier.
Thus, the basic idea is to identify
    new (possibly unlabeled) samples for which we cannot
    confidently predict the right label.
In standard approaches, a new sample is either 
    selected from a list of unlabeled samples 
    or generated based on:
    its high entropy over the classes,
    its large contribution to the loss function,
    or because it results in disagreement among different classifiers.
Clearly, when a sample is selected among unlabeled samples,
     it is important to label the sample correctly (e.g., using
     a human annotator).
The classifier is then retrained  
     with a larger dataset that contains samples
     that were hard to predict.
     
Alternatively, in adversarial learning, we generate new
     samples by perturbing correctly classified samples
     so as to have the system give the wrong output.
Thus, the idea here is that a small perturbation should not
     have resulted in a different classification.
Hence, by retraining the classifier with the old label,
     we expect the classifier to become more robust,
     and able to survive adversarial attacks.

Active learning is an iterative process.
After retraining, the process can be repeated to select 
      a new set of samples for the next iteration.
Clearly, the process can be stopped when
      no new samples can be generated or we
      are satisfied with the performance of the classifier.

In \cite{rana2023hybrid}, the authors consider
    a hybrid approach for reducing the number of frames
    and the number of video segment annotation samples
    for training classifiers on the UCF-101-24 and J-HMDB-21 datasets.
In the results, they show that annotating 5\% of the video frames
    can yield the same results as what can achieved with
    annotating 90\% of the frames.
For sample selection, the authors use a clustering
    method based on sample informativeness
    and diversity measures, and 
    a spatio-temporal weighted loss function.

In \cite{yang2003automatically}, the authors present
   different methods for selecting sample video frames from
   a long-term surveillance video from a geriatric care center.
Their goal was to identify several people in the video while
   minimizing the number of frames that need to be annotated.
Their approach was to develop different sample frame selection
   strategies and compare their performances based on
   their impact on the classification error.
Thus, the most effective sample selection strategies
    resulted in the rapid reduction of classification
    error through the human annotation of a small number
    of sample frames.
    
In \cite{goswami2023active}, the authors proposed
   an active learning approach in order to speed up the process
   of labeling digital video segments.
Their basic idea was to come up with a video
    summarization technique where the entire video is 
    replaced by a small number of video frames.
The video frames were selected based 
    on uncertainty and diversity measures computed
    over the video segment.
Thus, instead of requiring human annotators
    to review the entire video, they would only need
    to review the selected frames.

\subsection{Zero-shot learning} 
More recently, we have the recent introduction
   of zero-shot learning methods that have
   greatly benefited from the use of semantic
   information. 
Here, we use the term zero-shot learning
   to refer to methods that do not
   require any training (zero training)
   on specific video datasets associated
   with the task that needs to be learned.
These approaches can benefit from recent
   advancements in zero-shot
   learning in image analysis 
   (e.g., \cite{kirillov2023segment, estevam2021zero}).
Here, the basic idea is to
   use semantic information to come up
   with a textual description of an activity in
   terms of known categories.
For example, the activity of boiling
   an egg involves (i) detecting an egg,
   (ii) boiling the water, and (iii) placing
   the egg inside the water.
Thus, we can develop a zero-shot 
   method by combining systems
   that recognize an egg, boiling water, and
   placing an egg in water.
Clearly, the success of zero-shot
   learning depends on our ability
   to map semantic information into
   a collection of pre-trained classifiers
   or video activity recognition components
   that can closely approximate the given task.
We refer to \cite{estevam2021zero} for a survey
    of the different approaches.

\section{Concluding remarks}\label{sec:conclusion}
While the introduction of machine learning methods
   in video analysis has had a transformative impact,
   it has brought about several new challenges.
The standard use of supervised learning methods
   requires expensive labeling of large
   video datasets.
As a result, the majority of the methods
   for human activity recognition are trained
   and tested on relatively small datasets.
As an alternative, we have introduced
   low-parameter models that can be trained
   over candidate video segments. 
We also examined several methods
   for training with a limited number of labeled
   samples.
We believe that these methods hold great
   promise for the future.

Zero-shot learning eliminates
    the need for labeled samples.
However, it relies on the existence
    of pre-trained systems where
    new activities can be mapped to.
Active learning shows great promise
    because it minimizes the number of samples
    that need to be labeled while maximizing
    the classification performance.
Self-supervised learning attempts to learn
    the structure of the data for later adaptation.
Semi-supervised learning holds great promise in
    labeling a large number of samples and retraining
    based on them.
Overall, we expect significant
    growth in the application of these approaches
    to video analysis.
    
\section{Acknowledgement}
This material is based upon work supported by the National Science Foundation
under Grant No. 1613637, No. 1842220, and No. 1949230. Any opinions or findings
of this document reflect the author’s views, and they do not necessarily reflect the views
of NSF.

\bibliographystyle{IEEEtran}
\bibliography{references.bib}

\end{document}